\PassOptionsToPackage{hyphens}{url}
\documentclass{bytedance_seed}
\usepackage{amsmath,amsfonts,bm}

\def\eqref#1{equation~\ref{#1}}
\def\1{\bm{1}}

\DeclareMathAlphabet{\mathsfit}{\encodingdefault}{\sfdefault}{m}{sl}
\SetMathAlphabet{\mathsfit}{bold}{\encodingdefault}{\sfdefault}{bx}{n}

\usepackage{amsmath,amssymb}
\newcommand{\Rbf}{R_{\mathrm{bf16}}}
\newcommand{\tokenkl}{\widehat D_{\mathrm{KL}}}

\title{Where Rounding Happens Matters: Rounding-Location Alignment Stabilizes LLM Reinforcement Learning}
\author[1,2]{Yijing Ke}
\author[1,\dagger]{Tianle Cai}
\author[2,\dagger]{Ruosong Wang}
\affiliation[1]{ByteDance Seed}
\affiliation[2]{Peking University}
\contribution[\dagger]{\textsf{Corresponding authors}}
\date{\today}
\correspondence{\footnotesize Tianle Cai at \email{caitianle@bytedance.com} and Ruosong Wang at \email{ruosongwang@pku.edu.cn}}
\hypersetup{pdfauthor={Yijing Ke, Tianle Cai, Ruosong Wang}}
\abstract{Reinforcement learning for large language models often uses separate systems
for inference and training.  Even with identical weights and BF16 tensors,
implementation differences can make an intended on-policy update off-policy
and destabilize training.  Existing remedies align the entire inference and
training stacks, change global precision, or modify the learning algorithm.
We ask whether aligning selected parts of the execution stack can stabilize
BF16 RL while other implementation differences remain.  We align training-side
rounding locations with inference in three regions---RoPE, SwiGLU, and residual
RMSNorm---by retaining selected intermediates in FP32 until their BF16 outputs.
Across four Qwen-family models, aligning all three regions prevents the collapse
observed under the default BF16 baseline; RoPE-path alignment alone also
suffices in the studied runs, despite remaining numerical discrepancies.
The interventions keep the inference engine, matrix-multiplication and attention
kernels, and learning objective unchanged.
}
\begin{document}

\maketitle

\section{Introduction}

Reinforcement learning (RL) has become a central method for post-training large
language models (LLMs), improving mathematical reasoning, code generation, and
search-assisted problem solving
~\citep{shao2024deepseekmath,deepseek2025r1,jin2025searchr1}.
Maintaining stable optimization remains difficult, however, and model
performance can collapse during training
~\citep{yu2025dapo,zheng2025stabilizingrl}.
Training--inference mismatch is one source of this instability
~\citep{yao2025rollout,zhong2026diagnosing}.

Many RL systems use an inference engine such as vLLM to generate responses and
a separate implementation, such as Hugging Face Transformers, to recompute
their probabilities for training~\citep{kwon2023vllm,wolf2020transformers}.
Even with the same checkpoint and nominal data type, implementation differences
can make the recomputed probabilities disagree with those of the policy that
generated the responses.  This discrepancy turns an intended on-policy update
off-policy and can destabilize training
~\citep{yao2025rollout,zhong2026diagnosing,qi2025fp16}.

Prior studies stabilize training under this mismatch through global precision
changes~\citep{qi2025fp16} or algorithmic corrections
~\citep{yao2025rollout,zhong2026diagnosing}.
Another approach seeks exact numerical agreement by aligning arithmetic and
reduction order across engines~\citep{he2025nondeterminism,zhang2025tbik},
including the precision of intermediate computations in RoPE and
SwiGLU~\citep{panda2026mismatch}.
We ask whether aligning only selected parts of the execution stack can stabilize
BF16 RL while retaining the learning objective and allowing other implementation
differences to remain.

A compiled/eager comparison motivates this question.  Both compiled and eager
execution support stable training when inference and training use aligned modes;
the two crossed conditions collapse.  This observation directs attention to
agreement between the engines and motivates inspecting the numerical changes
introduced by fusion.
Separate kernels store some intermediates in BF16, whereas a fused kernel can
retain them in FP32 until its output.  We call this difference in where values
are rounded a \emph{rounding-location mismatch}.

Guided by graph inspection, we align training-side rounding locations with
those in inference in three regions: the RoPE path~\citep{su2021roformer},
SwiGLU~\citep{shazeer2020glu}, and residual RMSNorm~\citep{zhang2019rmsnorm}.
We retain selected training intermediates in FP32 and round each region's
outputs to BF16, preserving BF16 model weights, the inference engine,
matrix-multiplication and attention kernels, and learning objective.

\paragraph{Contribution.}
We show that selective rounding-location alignment can stabilize BF16 LLM RL
without full-stack numerical agreement or additional mismatch corrections in
the learning objective.  Across four Qwen-family models, aligning all three
regions prevents the collapse observed under the default BF16 baseline despite
numerical disagreement; RoPE-path alignment alone also suffices in every studied
run.  These results establish selective alignment as a practical route to
stable training in the studied settings.

\section{Related Work}
\label{sec:related-work}

\paragraph{Corrections for training--inference mismatch.}
Truncated importance sampling and sequence-level trust-region masking limit the
effect of distribution shift between inference and training policies
~\citep{yao2025rollout,li2025trm}.
\citet{zheng2025stabilizingrl} analyze how numerical disagreement and policy
staleness affect token-level surrogate objectives.  Other remedies reduce the
learning rate when response length signals impending instability
~\citep{zhang2026beyondprecision}, or accept candidate updates using an
inference-side discrepancy proxy~\citep{liang2026mipu}.

\paragraph{Execution alignment.}
Batch-invariant kernels fix reduction structure to remove numerical dependence
on batch size~\citep{he2025nondeterminism}; invariant reductions can also extend
agreement across tensor-parallel sizes~\citep{zhang2025tbik}.
\citet{zhong2026diagnosing} use the same Hugging Face implementation for inference
and training with batch-invariant kernels.  Their experiments show that numerical
disagreement alone can cause training collapse.  In mixture-of-experts models,
replaying the expert selections made during inference reduces routing
disagreement and stabilizes training~\citep{ma2025routing}.
The Ring-linear report aligns training and inference module by
module~\citep{ling2025everyattention}; \citet{panda2026mismatch} explicitly
matches rounding locations in RoPE and SwiGLU as part of achieving
bitwise-identical token log probabilities.

\paragraph{Precision and quantization.}
\citet{qi2025fp16} show that switching from BF16 to FP16 reduces
training--inference mismatch and improves RL stability in their experiments.
Low-bit RL systems also address the tradeoff between inference efficiency
and numerical agreement.  \citet{huang2025qerl} combine NVFP4 quantization with
low-rank adaptation to accelerate rollouts and study quantization noise as a
source of exploration.  Other approaches combine FP8 linear-layer
computation and KV-cache quantization with importance-sampling corrections
~\citep{qiu2026fp8rl}, or align the FP8 precision flow across training and
rollouts to reduce numerical mismatch~\citep{xi2026jetrl}.
QaRL aligns low-bit matrix multiplications in the training forward pass with
quantized inference, together with importance correction and a modified
policy objective~\citep{gu2026qarl}.
Related work on long-context training shows that BF16 errors can disrupt the
relative-position property of rotary position embedding (RoPE)
~\citep{wang2024precisionposition}.

\section{Training-Inference Mismatch in LLM RL}
\label{sec:training-inference-mismatch}

LLM RL commonly separates response generation from optimization.  An inference
engine such as vLLM generates responses token by token, using an implementation
optimized for efficient generation.  A training implementation such as Hugging
Face Transformers then processes those same responses to recompute their token
probabilities and differentiate the learning objective
~\citep{kwon2023vllm,wolf2020transformers}.  At identical weights, these two
implementations are intended to represent the same policy.  Numerical differences
can make their probabilities disagree before any optimization step, creating
\emph{training--inference mismatch} in addition to the policy changes caused by
learning.  This mismatch poses a major challenge for LLM RL, as it can
destabilize training and cause performance collapse
~\citep{yao2025rollout,zhong2026diagnosing,qi2025fp16}.

\subsection{Where the mismatch comes from}

The same model can produce different floating-point results in different
implementations.  We distinguish differences in precision and casts, kernels,
and reduction order.

\paragraph{Precision and casts.}
Parameter, buffer, computation, and accumulation precision can differ, as can
cast locations.  To accelerate generation, some inference engines quantize
the KV cache or use low-precision computation, reducing memory traffic or
arithmetic cost; these choices can introduce further disagreement with
higher-precision training~\citep{qiu2026fp8rl,xi2026jetrl}.
BF16 tensor storage also does not imply that every intermediate is computed in
BF16.  For two operations $f$ and $g$, let $\Rbf$ denote rounding to BF16.
Rounding the intermediate and retaining it in FP32 give, respectively,
\[
  \underbrace{\Rbf\!\left(g(\Rbf(f(x)))\right)}_{\text{intermediate rounded to BF16}}
  \qquad\text{and}\qquad
  \underbrace{\Rbf\!\left(g(f(x))\right)}_{\text{intermediate retained in FP32}} .
\]
Both return BF16 outputs, but $g$ can receive different values.  We call this a
\emph{rounding-location mismatch}.  Compilation can introduce this difference
through \emph{kernel fusion}, which combines operations in one kernel: a fused
kernel can retain an intermediate in FP32, whereas separate kernels may store
and reload it in BF16~\citep{ansel2024pytorch2}.

\paragraph{Kernels.}
The same operation can use different kernels or decompositions.  vLLM combines
Q, K, and V projections and gate and up projections, whereas Hugging Face may
execute them separately; the systems can
also use different attention implementations~\citep{dao2022flashattention}.

\paragraph{Reduction order.}
Floating-point addition is not associative, so batch size, tensor parallelism,
and kernel tiling can change a reduction's result.  Invariant kernels across
batch and parallel sizes fix the reduction structure.

\subsection{Algorithm and experimental setup}

We define the sampling process, policy update, and discrepancy measure used in
our experiments.

\paragraph{Policies and advantages.}
At the start of an iteration, let $\theta_0$ denote the checkpoint used for
inference.  The inference engine uses policy $\mu_{\theta_0}$ to sample $G$
responses $y_i=(a_{i,1},\ldots,a_{i,T_i})$ for a prompt
$x\sim\mathcal D$.  The training engine starts from $\theta=\theta_0$ and uses
policy $\pi_\theta$ to recompute the token probabilities of those responses.
Even at identical weights, the policies can differ numerically.
Each response receives a correctness reward $R_i$, and its advantage is centered
within the group without standard-deviation normalization:
\begin{equation}
  A_i=R_i-\bar R,
  \qquad \bar R=\frac{1}{G}\sum_{j=1}^{G}R_j .
  \label{eq:group-advantage}
\end{equation}

\paragraph{Objective.}
Prior work addresses inference--training mismatch with importance-sampling
corrections, truncation, and masking rules
~\citep{yao2025rollout,li2025trm,zheng2025stabilizingrl}.
To study execution alignment under a simple learning objective, we use
PPO-clipped GRPO~\citep{schulman2017ppo,shao2024deepseekmath} without additional
mismatch corrections, and hold this objective fixed across conditions.  Let
$s_{i,t}=(x,a_{i,<t})$ and define the training-side ratio
$r_{i,t}(\theta)=\frac{\pi_\theta(a_{i,t}\mid s_{i,t})}
{\pi_{\theta_0}(a_{i,t}\mid s_{i,t})}$.  With expectation over the sampling
process above, the update minimizes
\begin{equation}
  \mathcal L_{\mathrm{GRPO}}(\theta)
  =-\mathbb E_{\mu_{\theta_0}}\Biggl[\frac{1}{G}\sum_{i=1}^{G}\sum_{t=1}^{T_i}
    \min\Bigl\{r_{i,t}(\theta)A_i,
    \operatorname{clip}\!\left(r_{i,t}(\theta),
      1-\epsilon_{\mathrm{low}},1+\epsilon_{\mathrm{high}}\right)A_i
    \Bigr\}\Biggr].
  \label{eq:grpo-loss}
\end{equation}
The denominator, sampled responses, and advantages are fixed; gradients flow
through $\pi_\theta$.  At the start of an update, $r_{i,t}(\theta_0)=1$ because
both probabilities use the training implementation, even if it disagrees with
the inference policy that sampled the responses.  Clipping therefore controls
training-policy changes while leaving the initial discrepancy between
$\mu_{\theta_0}$ and $\pi_{\theta_0}$ uncorrected.

\paragraph{Mismatch measurement.}
To measure numerical disagreement, we compare the two implementations at the
same fixed parameters $\theta_0$.  For tokens sampled from $\mu_{\theta_0}$,
we use the following estimate of per-token forward Kullback--Leibler (KL)
divergence~\citep{kullback1951information}:
\begin{equation}
  \tokenkl
  =\frac{1}{N}\sum_{i=1}^{G}\sum_{t=1}^{T_i}
   \left[\log\mu_{\theta_0}(a_{i,t}\mid s_{i,t})
        -\log\pi_{\theta_0}(a_{i,t}\mid s_{i,t})\right],
  \qquad N=\sum_{i=1}^{G}T_i .
  \label{eq:per-token-kl}
\end{equation}

\paragraph{Experimental setup.}
The main-text RL experiments use Sanity-Test-R1D-1.5B, a MATH subset selected to be neither
trivial nor unsolvable for the initial model~\citep{hendrycks2021math,qi2025fp16},
and evaluate on problems from the 2025 American Invitational Mathematics Examination
(AIME 2025)~\citep{maa2026aime}.  vLLM generates responses; Hugging Face Transformers
recomputes their probabilities and updates the model, with fully sharded data
parallelism (FSDP)~\citep{zhao2023fsdp} as the distributed sharding strategy.
Appendix~\ref{app:experimental-details} reports execution settings, sample counts,
and optimization hyperparameters.

\section{Motivation: Small Execution Changes Can Destabilize RL}
\label{sec:execution-mode}

We motivate the study with a small change to execution: enabling or disabling
compilation during inference and training.  We use
DeepSeek-R1-Distill-Qwen-1.5B in BF16, following the experimental setup in
Section~\ref{sec:training-inference-mismatch}.

We compare compiled and eager execution on the vLLM inference side and the
Hugging Face training side.  The four conditions are
compiled--compiled, compiled--eager, eager--compiled, and eager--eager, where
the first mode denotes inference.  In the two aligned conditions, inference and
training use the same execution mode; in the two crossed conditions, they use
different modes.  Appendix~\ref{app:experimental-details} gives the exact execution
settings.

\begin{figure}[ht]
    \centering
    \includegraphics{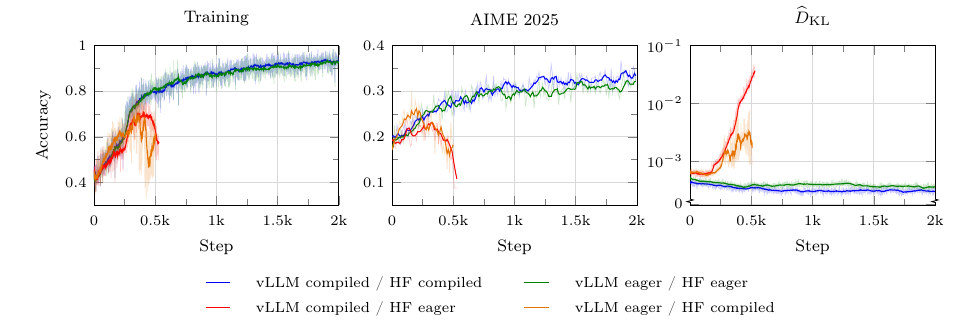}
    \caption{Training accuracy, AIME 2025 accuracy, and the inference--training
    discrepancy $\tokenkl$ (Eq.~\ref{eq:per-token-kl}) for the four execution-mode
    conditions.  Observations are smoothed using causal trailing means over
    20 observations for training accuracy and $\tokenkl$ and five observations
    for AIME 2025 accuracy.
    Both aligned conditions maintain high accuracy; both crossed
    conditions initially improve and then lose accuracy.}
    \label{fig:reward-curves}
\end{figure}
Figure~\ref{fig:reward-curves} shows that both compiled and eager execution
support stable training when inference and training use aligned modes.  The
two crossed conditions collapse, accompanied by rising $\tokenkl$, whereas
both aligned conditions retain a small, nonzero discrepancy.

With eager training held fixed, changing inference from compiled to eager
restores stability without increasing training-side precision.  Together, the
four conditions support agreement between execution modes as a stabilizing
factor.  They motivate inspecting the numerical changes introduced by
compilation to identify rounding locations for targeted alignment.

\section{Qwen Analysis: Where Fusion Moves Rounding Locations}
\label{sec:qwen-kernel-fusion}

We identify rounding-location mismatches by inspecting intermediate data types
and fused operations in the inference and training graphs.  The aim
is to find values stored in BF16 in one implementation but retained in FP32
until a later operation in the other.

For Qwen2 and Qwen3~\citep{yang2024qwen2,yang2025qwen3}, inspection of the
compiled and eager vLLM paths identifies three regions in which fusion changes
rounding locations: Q/K normalization, when present, plus RoPE; SwiGLU's
sigmoid linear unit (SiLU) activation~\citep{elfwing2018silu} and gate
multiplication; and residual addition plus root mean square normalization
(RMSNorm).

\begin{figure}[ht]
    \centering
    \includegraphics{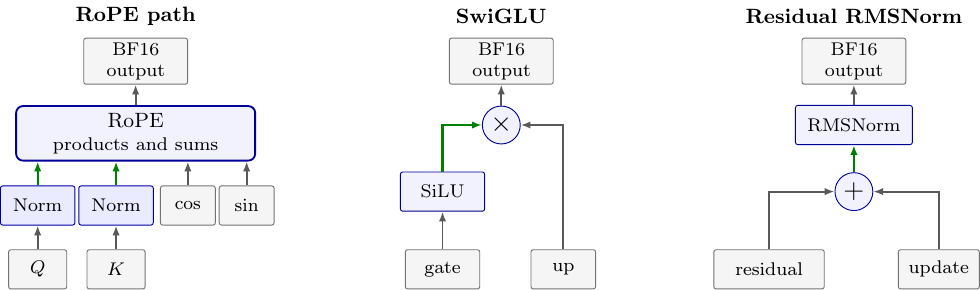}
  \caption{The three Qwen regions in which fusion changes rounding locations.
  Green arrows carry intermediates retained in FP32 under fusion and rounded
  to BF16 between separate operations.  RoPE products and sums are also retained
  in FP32 under fusion.  Region outputs are BF16 in both cases.  Q/K
  normalization, when present, is part of the RoPE path.}
  \label{fig:qwen-rounding-locations}
\end{figure}

Figure~\ref{fig:qwen-rounding-locations} shows the identified locations.  Fused
execution retains the marked intermediates in FP32 and stores region outputs
in BF16.  Separate operations round those intermediates to BF16 before passing
them to the next operation.

\subsection{The three affected regions}

\paragraph{Q/K normalization and RoPE.}

In variants with Q/K normalization, fusion retains its statistics, normalized
and scaled Q/K values, and subsequent RoPE products and sums in FP32.  Variants
without Q/K normalization still fuse RoPE products and sums.  Only rotated Q/K
outputs are stored in BF16.

\paragraph{SwiGLU.}

Fusion retains sigmoid and SiLU results in FP32 through multiplication with the
up projection, then stores the product in BF16.  Eager execution stores the SiLU
output in BF16 before multiplication.

\paragraph{Residual RMSNorm.}

Under fusion, RMSNorm consumes the unrounded FP32 residual sum and retains
its intermediate computations in FP32.  Both its output and a separate copy
of the sum for the continuing residual stream are stored in BF16.  Eager
execution rounds the sum before RMSNorm.  This region follows both attention
and MLP in every decoder block.

The intervention retains the identified training intermediates in FP32 and
casts each region output back to BF16.  Changes are confined to pointwise and
normalization operations in the RoPE, SwiGLU, and residual-RMSNorm regions,
preserving the inference engine and the compute-intensive matrix-multiplication
and attention kernels.  Aligning all three regions adds 5.33--10.92\% to
forward-and-backward latency, including communication, and 0.41--0.89~GiB of
peak allocated memory per GPU across three benchmarked models
(Section~\ref{app:training-efficiency}).  The additional training time is
expected to represent a smaller percentage of total RL time once rollout
generation is included.

\section{Static Attribution of Policy Discrepancy}
\label{sec:fp32-region-ablation}

We quantify how aligning the three regions identified in
Section~\ref{sec:qwen-kernel-fusion} changes policy discrepancy at fixed model
weights.  In Hugging Face recomputation, the interventions keep in FP32,
respectively: Q/K through RoPE (including normalization when present); SiLU output and
gate multiplication; and residual addition and RMSNorm.

Each region output is cast back to BF16 at its original location; model weights,
matrix multiplications, and attention kernels are unchanged.  We compute all
$2^3$ combinations on fixed inference responses and log probabilities; only
Hugging Face recomputation changes.  For simplicity, the static analyses here
and in Section~\ref{sec:bf16-cell-phase-fourier} use three models, omitting
Qwen3-4B-Base because its architecture closely resembles Qwen3-8B-Base.
Table~\ref{tab:fp32-regions} reports the estimator in
\eqref{eq:per-token-kl}.  Appendix~\ref{app:experimental-details}
gives the prompt, response, and token counts.

\begin{table}[!htb]
  \caption{Estimated per-token KL divergence from vLLM to Hugging Face.
  Check marks indicate internal FP32 computation during recomputation; bold
  values are column minima.  The first eight rows report $\tokenkl\times10^4$,
  and the final two give percentage reductions from the BF16 baseline.}
  \label{tab:fp32-regions}
  \begin{center}
  \small
  \setlength{\tabcolsep}{3pt}
  \begin{tabular}{ccc rrr}
    \toprule
    \multicolumn{3}{c}{Region computed in FP32} &
    \multicolumn{3}{c}{$\tokenkl\times 10^4$} \\
    \cmidrule(lr){1-3}\cmidrule(lr){4-6}
    \shortstack{RoPE path} & SwiGLU & Residual RMSNorm &
    \shortstack{DeepSeek-R1-Distill-\\Qwen-1.5B} &
    \shortstack{Qwen3-1.7B} &
    \shortstack{Qwen3-8B-Base} \\
    \midrule
    & & & 6.5416 & 7.4461 & 3.7599 \\
    $\checkmark$ & & & 5.0951 & 7.1912 & 3.4247 \\
    & $\checkmark$ & & 6.1650 & 7.6227 & 4.1681 \\
    & & $\checkmark$ & 6.1040 & 7.4906 & 3.4921 \\
    $\checkmark$ & $\checkmark$ & & 5.0548 & 6.9537 & 3.4139 \\
    $\checkmark$ & & $\checkmark$ & 5.0716 & 6.8715 & 3.6027 \\
    & $\checkmark$ & $\checkmark$ & 6.4336 & 7.7358 & 3.7176 \\
    $\checkmark$ & $\checkmark$ & $\checkmark$ &
    \textbf{4.8741} & \textbf{6.7689} & \textbf{3.2891} \\
    \midrule
    \multicolumn{3}{l}{RoPE-path reduction (\%)} & 22.1 & 3.4 & 8.9 \\
    \multicolumn{3}{l}{Three-region reduction (\%)} & 25.5 & 9.1 & 12.5 \\
    \bottomrule
  \end{tabular}
  \end{center}
\end{table}
Computing all three regions in FP32 yields the smallest estimate for every
model.  Relative to the all-BF16 condition, $\tokenkl$ decreases by 25.5\%,
9.1\%, and 12.5\% for the three models, respectively.

We use Shapley attribution~\citep{shapley1953value} to allocate the joint KL
reduction among the three interventions, averaging each intervention's marginal
contribution over all orders in which they can be applied
(Appendix~\ref{app:shapley-attribution}).  The RoPE path accounts for 81.5\%,
91.8\%, and 76.9\% of the joint reduction.  The SwiGLU shares are 8.6\%,
$-3.8$\%, and $-14.3$\%; the residual-RMSNorm shares are 9.9\%, 12.0\%, and 37.4\%.

The RoPE path is the largest contributor to the joint discrepancy reduction
in all three models.  The negative SwiGLU shares for Qwen3-1.7B and Qwen3-8B-Base
mean that adding this intervention increases discrepancy on average across
intervention orders.  The nonmonotonic results
show that higher intermediate precision does not automatically reduce mismatch.
We next evaluate these interventions in RL training.

\section{Rounding-Location Alignment Stabilizes Training}
\label{sec:llm-rl-controls}

The static attribution motivates testing both RoPE-path alignment and alignment
of all three regions.  We now evaluate whether these interventions stabilize
RL training.  We use the objective and data from
Section~\ref{sec:training-inference-mismatch} with four Qwen-family models
spanning Qwen2 and Qwen3: DeepSeek-R1-Distill-Qwen-1.5B,
Qwen3-1.7B, Qwen3-4B-Base, and Qwen3-8B-Base.
Appendix~\ref{app:experimental-details} gives the sampling, optimization, and
execution settings.  We compare four conditions for each model:

\begin{itemize}
    \item \textbf{BF16 baseline:} default vLLM inference and Hugging Face training.
    \item \textbf{FP32 RoPE path:} only the training-side RoPE path retains FP32
          intermediates, including Q/K normalization when present.
    \item \textbf{FP32 at all three regions:} the training-side RoPE, SwiGLU,
          and residual-RMSNorm regions retain FP32 intermediates.
    \item \textbf{Zero-mismatch backend:} the same Hugging Face model
          implementation is used for inference and
          training, with batch-invariant kernels.  This aligns precision and
          casts, kernels, and reduction order.  Section~\ref{app:zero-mismatch-backend}
          describes the implementation.
\end{itemize}

\begin{figure}[!t]
    \centering
    \includegraphics{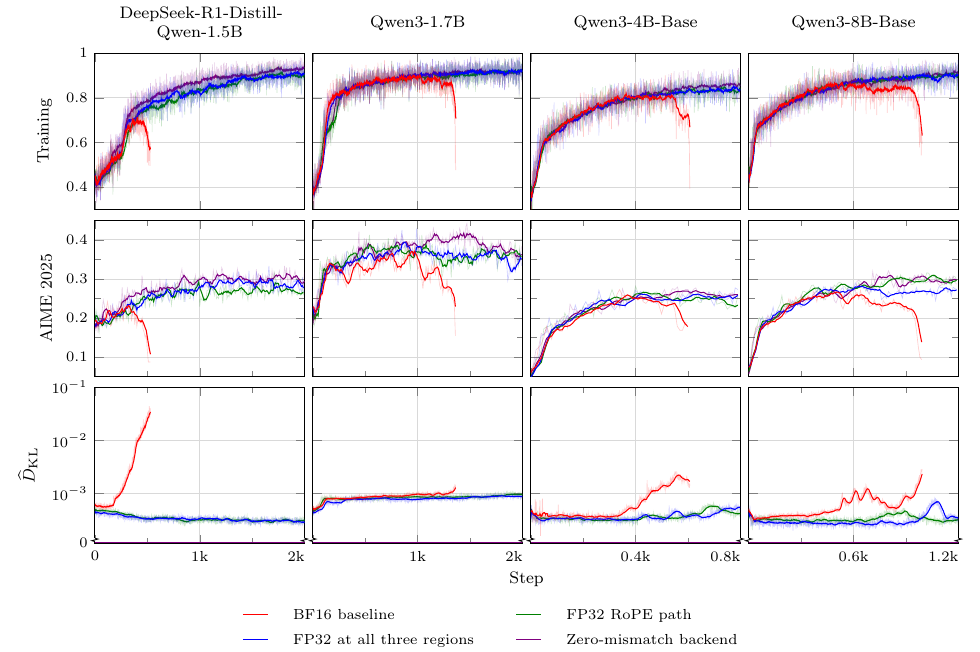}
    \caption{Training accuracy (top), AIME 2025 accuracy (middle), and
    inference--training discrepancy $\tokenkl$ (bottom; Eq.~\ref{eq:per-token-kl}).
    Observations are smoothed using causal trailing means over
    20 observations for training accuracy and $\tokenkl$ and five observations
    for AIME 2025 accuracy.
    The BF16 baseline collapses for every model; the FP32 RoPE path, FP32 at
    all three regions, and Zero-mismatch backend remain stable.}
    \label{fig:llm-rl-model-reward-curves}
\end{figure}
Figure~\ref{fig:llm-rl-model-reward-curves} shows that the BF16 baseline improves
and then loses accuracy for every model.  Both targeted alignments and the
Zero-mismatch backend avoid collapse and retain higher AIME 2025 accuracy.
$\tokenkl$ rises as the baseline collapses, sharply for DeepSeek and more
modestly for Qwen3-1.7B.  Both targeted alignments retain small, nonzero
$\tokenkl$, while the Zero-mismatch backend remains at zero.
These results show that inspecting the graphs and aligning the identified
rounding locations greatly stabilizes BF16 RL.  Retaining FP32 intermediates
along the RoPE path alone is sufficient in all observed runs.

The Zero-mismatch backend confirms that the same objective can train stably
with aligned execution.  The two targeted conditions show that eliminating
all numerical disagreement is unnecessary for stability in these runs:
aligning the RoPE path alone suffices, with no additional correction in the
learning objective.  This is evidence about training stability, complementing
the fixed-weight discrepancy reductions in Section~\ref{sec:fp32-region-ablation}.
Appendix~\ref{app:dapo-math} repeats these controls with
DeepSeek-R1-Distill-Qwen-1.5B on DAPO-Math-17k~\citep{yu2025dapo} and likewise
finds that RoPE-path alignment stabilizes training.

RoPE-path alignment also has lower overhead than aligning all three regions:
it adds only 0.90--2.04\% to training-computation latency, with unchanged peak
allocated memory, in the three-model benchmark
(Section~\ref{app:training-efficiency}).

\section{Structure of BF16 Rounding Phases}
\label{sec:bf16-cell-phase-fourier}

We explore the numerical structure of intermediates at the identified rounding
locations.  This characterization complements the policy-discrepancy and RL
experiments by examining the values directly affected by rounding.
A value's \emph{BF16 cell phase} is its normalized position between adjacent
BF16 values; away from midpoint ties, this position determines the direction
and relative size of rounding error.  Using the same three models as in
Section~\ref{sec:fp32-region-ablation}, we compare RoPE products, SiLU outputs,
and residual sums with projection, attention, and RMSNorm outputs.

For each signal $s$ and layer $\ell$, the phase-concentration statistic
$A_{\ell,s}$ is the root mean square of the first 16 Fourier magnitudes of
the phase distribution; larger values indicate stronger departures from
uniformity over these frequencies.  We report its median across layers,
$\widehat A_s$.  Section~\ref{app:cell-phase-fourier} gives the sampling
procedure, formal definitions, and individual frequency spectra.

\begin{figure}[!t]
  \centering
  \includegraphics{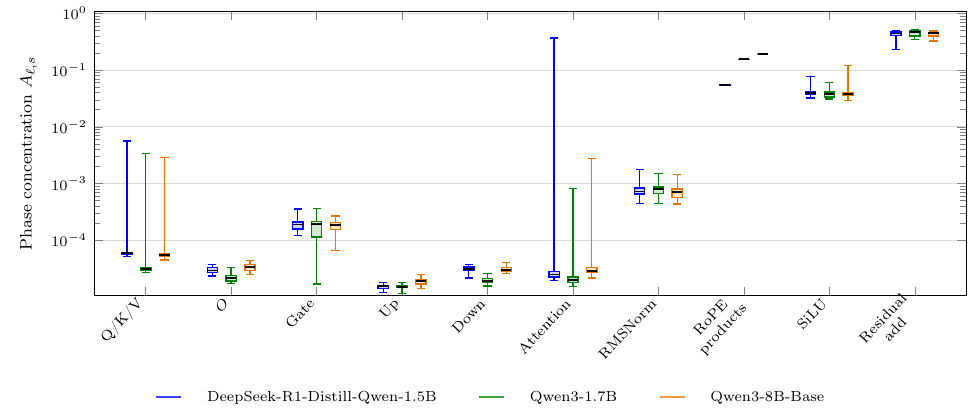}
  \caption{Layer distributions of the 16-frequency phase concentration.  The
  center line is the layer median, boxes span the first and third quartiles,
  and whiskers show the full range.  The three intermediates at changed
  rounding locations are markedly more nonuniform than the other signals.}
  \label{fig:cell-phase-overall}
\end{figure}

Figure~\ref{fig:cell-phase-overall} compares the groups.  Across models,
$\widehat A_s$ is $1.5\times10^{-5}$--$8.0\times10^{-4}$ for
projection, attention, and RMSNorm outputs, compared with 0.038--0.040 for
SiLU, 0.054--0.192 for RoPE products, and 0.443--0.472 for residual sums.  The
three groups whose rounding locations change under fusion have the largest
median phase concentrations.  Some Q/K/V and attention layers have large
maxima, but low medians and quartiles show that they are isolated.

The three regions identified by graph inspection thus exhibit distinctive
BF16 cell-phase structure across all three models.  The individual spectra
also show consistent frequency patterns (Section~\ref{app:cell-phase-fourier}).
These measurements quantify how values are distributed within BF16 intervals;
their rounding errors and downstream effects additionally depend on interval
widths and subsequent computations (Section~\ref{app:rounding-error}).
The Fourier analysis provides a complementary view of the affected
intermediates alongside the static discrepancy reductions and RL stabilization
reported in Sections~\ref{sec:fp32-region-ablation} and~\ref{sec:llm-rl-controls}.

\section{Discussion}

\paragraph{Why a small gap matters.}
Aligning all three regions reduces the static per-token KL by 9.1--25.5\%,
yet prevents collapse in the RL runs.  One possible explanation is that
numerical errors at different locations affect training updates differently.
Aligning a sensitive region may therefore improve stability substantially
while leaving much of the average policy discrepancy intact.

\paragraph{Why not simply compile training?}
Compilation is a possible remedy: the compiled--compiled condition is stable
in the motivating experiment.  It does, however, introduce practical costs.
Hugging Face training compilation covers both forward and backward passes and
incurs a substantial initial compilation cost
~\citep{huggingface2026torchcompile}.  Variable response lengths can trigger
recompilation, while data-dependent control flow and unsupported operations
can cause graph breaks~\citep{pytorch2026compiletroubleshooting}.  Distributed
sharding and custom kernels may also require compiler-specific integration.
A reported large-scale training deployment required explicit work on graph
breaks, custom kernels, and numerical agreement with eager execution
~\citep{pytorch2024fsdpcompile}.  Compiling Hugging Face also does not make it
the same implementation as compiled vLLM: precision and casts, kernels, and
reduction order can still differ.

\paragraph{Alignment or increased training-side precision?}
With the eager training implementation and its precision held fixed, switching
inference from compiled to eager execution restores stability
(Section~\ref{sec:execution-mode}).  Together with the crossed execution-mode
controls, this supports execution alignment as a stabilizing factor and shows
that stability can be restored without increasing training-side precision.
Section~\ref{sec:fp32-region-ablation} also shows that higher intermediate
precision does not automatically reduce static policy discrepancy.
Together with the targeted rounding-location interventions, these results
support rounding-location alignment as a stabilizing mechanism.

\paragraph{Limitations.}
The experiments cover two math datasets and models from a single architectural
family.  Extending the intervention to other architectures requires inspecting
their inference and training graphs to identify the relevant rounding locations
and adapt the alignment accordingly.  The results demonstrate the importance of
rounding-location alignment for stable LLM RL; broader experiments are needed
to establish when it is sufficient across architectures and training settings.

\paragraph{Takeaway.}
The results support inspecting the inference and training graphs to
identify and align rounding locations moved by fusion as a targeted intervention
for stabilizing BF16 RL.  RoPE-path alignment alone suffices to
stabilize the studied BF16 RL runs.  Aligning all three regions stabilizes
training and gives the lowest per-token KL among the conditions compared in
Section~\ref{sec:fp32-region-ablation}.  The interventions keep the inference
engine, matrix-multiplication and attention kernels, and training objective
unchanged.  Stable training therefore does not require full-stack alignment
in these experiments.

\clearpage
\bibliographystyle{plainnat}
\bibliography{references}

\clearpage
\beginappendix
\section{Experimental Details}
\label{app:experimental-details}

The main-text experiments use the Sanity-Test-R1D-1.5B training
set and evaluate on AIME 2025.  Unless a condition states
otherwise, vLLM and Hugging Face use their default configurations: inference uses vLLM with FlashAttention,
and Hugging Face training uses scaled dot-product attention without
compilation.  All experiments use the default sampling configuration:
temperature $1.0$, with no top-$p$ or top-$k$ filtering.  Responses are
truncated when they reach the configured length limit.

\subsection{Static analyses}

Section~\ref{sec:fp32-region-ablation} uses the first 512 prompts, and
Section~\ref{sec:bf16-cell-phase-fourier} uses the first 256.  In both analyses,
we generate one response per prompt.  Section~\ref{sec:fp32-region-ablation} holds the responses
and inference log probabilities fixed across all eight Hugging Face
recomputations.  Table~\ref{tab:static-analysis-details} reports the sample
sizes; token counts include generated response tokens only.

\begin{table}[ht]
  \caption{Sampling configuration for Sections~\ref{sec:fp32-region-ablation}
  and~\ref{sec:bf16-cell-phase-fourier}.  Each section uses one response per
  prompt, so the prompt and response counts are equal.}
  \label{tab:static-analysis-details}
  \begin{center}
  \small
  \setlength{\tabcolsep}{3.5pt}
  \begin{tabular}{lrrr}
    \toprule
    Configuration & DeepSeek-R1-Distill-Qwen-1.5B & Qwen3-1.7B & Qwen3-8B-Base \\
    \midrule
    \multicolumn{4}{l}{\textit{Section~\ref{sec:fp32-region-ablation}}} \\
    Prompts & 512 & 512 & 512 \\
    Responses & 512 & 512 & 512 \\
    Response tokens & 2,860,411 & 2,615,319 & 548,171 \\
    Response-token limit & 10,240 & 5,120 & 5,120 \\
    Model-length limit & 16,384 & 16,384 & 16,384 \\
    \midrule
    \multicolumn{4}{l}{\textit{Section~\ref{sec:bf16-cell-phase-fourier}}} \\
    Prompts & 256 & 256 & 256 \\
    Responses & 256 & 256 & 256 \\
    Response tokens & 1,050,873 & 1,308,191 & 270,510 \\
    Response-token limit & 5,120 & 5,120 & 5,120 \\
    Model-length limit & 16,384 & 16,384 & 16,384 \\
    \bottomrule
  \end{tabular}
  \end{center}
\end{table}

\subsection{RL training and evaluation}

Table~\ref{tab:rl-training-details} summarizes the RL configuration used in
Sections~\ref{sec:execution-mode} and~\ref{sec:llm-rl-controls}.  Training
metrics are logged once per step, and AIME accuracy is evaluated every 10
steps.  At each AIME evaluation, every GPU independently
generates one response for each question in the full test set.  We compute
accuracy on each GPU and report the mean across all GPUs.

\begin{table}[ht]
  \caption{RL training and evaluation configuration.}
  \label{tab:rl-training-details}
  \begin{center}
  \small
  \setlength{\tabcolsep}{3pt}
  \begin{tabular}{lrrrr}
    \toprule
    Configuration & \shortstack{DeepSeek-R1-Distill-\\Qwen-1.5B} &
    \shortstack{Qwen3-1.7B} & \shortstack{Qwen3-4B-\\Base} &
    \shortstack{Qwen3-8B-\\Base} \\
    \midrule
    GPUs & 8 & 8 & 16 & 32 \\
    Prompts per step & 64 & 64 & 64 & 64 \\
    Responses per prompt & 8 & 8 & 8 & 8 \\
    Response-token limit & 10,240 & 10,240 & 10,240 & 10,240 \\
    Evaluation model-length limit & 16,384 & 16,384 & 16,384 & 16,384 \\
    Mini-batches per step & 4 & 1 & 1 & 1 \\
    \midrule
    Optimizer & \multicolumn{4}{c}{Adam~\citep{kingma2015adam} $(\beta_1,\beta_2)=(0.9,0.95)$} \\
    Learning rate & \multicolumn{4}{c}{$10^{-6}$} \\
    PPO clipping $(\epsilon_{\mathrm{low}},\epsilon_{\mathrm{high}})$ &
    \multicolumn{4}{c}{$(0.2,0.28)$} \\
    \bottomrule
  \end{tabular}
  \end{center}
\end{table}

For Section~\ref{sec:execution-mode}, compiled vLLM uses its default path, with
\texttt{enforce\_eager} set to \texttt{False}; eager vLLM sets it to
\texttt{True}.  Compiled Hugging Face training applies
\texttt{torch.compile()} to every decoder layer; eager training executes the
model directly.

\subsection{Static attribution}
\label{app:shapley-attribution}

For Section~\ref{sec:fp32-region-ablation}, let $C$ be the set of three
interventions and $D(S)$ the value of $\tokenkl$ with subset $S$ enabled.
The Shapley contribution of region $j$ averages its marginal reduction over
all intervention orders:
\begin{equation}
  \Delta_j=\sum_{S\subseteq C\setminus\{j\}}
  \frac{|S|!(|C|-|S|-1)!}{|C|!}
  \left[D(S)-D(S\cup\{j\})\right].
  \label{eq:shapley-reduction}
\end{equation}
Normalizing by the joint reduction gives shares
$\Delta_j/[D(\varnothing)-D(C)]$ that sum to one.  Positive shares represent
average decreases in discrepancy across intervention orders; negative shares
represent average increases.

The token log-probability differences in Section~\ref{sec:fp32-region-ablation}
can also be summed within each response.  For the baseline conditions, the mean of these
response-level log-probability differences is 3.65, 3.80, and 0.40 nats for
DeepSeek-R1-Distill-Qwen-1.5B, Qwen3-1.7B, and Qwen3-8B-Base, respectively.
These values equal $\tokenkl$ multiplied by the mean response length.  They
illustrate how small per-token differences accumulate over a response.

\subsection{BF16 cell phases and Fourier diagnostics}
\label{app:cell-phase-fourier}

For Section~\ref{sec:bf16-cell-phase-fourier}, a \emph{signal} collects scalars
from an FP32 intermediate where eager execution would store it in BF16.
We compare ten groups: Q/K/V projection outputs; the attention output
projection ($O$); gate, up, and down projection outputs; attention output;
RMSNorm output; four RoPE products; SiLU output; and two residual sums per
decoder block.  Fusion changes rounding locations in the final three groups.
Each is computed in FP32 from operands observed during the BF16 forward pass.
Each scalar at a selected generated-token position is a sample; we analyze
layers separately and exclude prompt and padding positions.  All sampled
elements were finite.
Table~\ref{tab:static-analysis-details} reports the prompt, response, and token
counts for DeepSeek-R1-Distill-Qwen-1.5B, Qwen3-1.7B, and Qwen3-8B-Base.

For a nonzero, finite FP32 value $x$ in the normal range of BF16, let
$e(x)=\lfloor\log_2|x|\rfloor$.  BF16 has 7 explicit fraction bits (8-bit
significand precision)~\citep{kalamkar2019study}, so adjacent values within
$[2^{e(x)},2^{e(x)+1}]$ are spaced by $h(x)=2^{e(x)-7}$.  Define the cell phase
\begin{equation}
  \phi(x)=\operatorname{frac}(x/h(x)), \qquad 0\leq \phi(x)<1,
  \label{eq:cell-phase}
\end{equation}
where $\operatorname{frac}(z)=z-\lfloor z\rfloor$; set $\phi(0)=0$.

\begin{figure}[!t]
  \centering
  \includegraphics{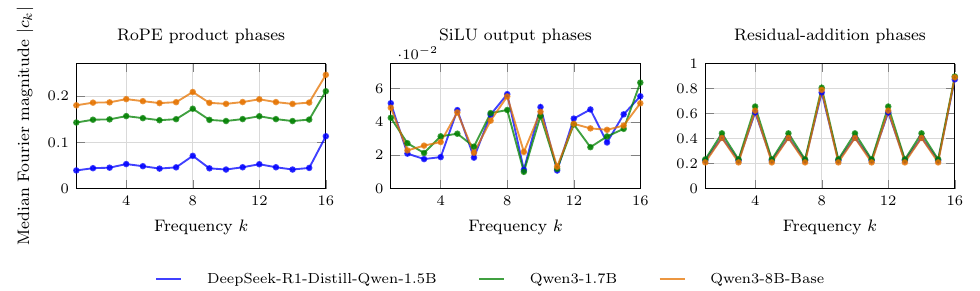}
  \caption{Fourier magnitudes for RoPE products, SiLU outputs, and residual
  sums.  Each point summarizes $|\widehat c_{\ell,s,k}|$ by its median across
  layers.  For residual addition, the magnitudes of the attention and MLP
  residual sums are averaged within each layer before taking the median.
  The panels use different vertical scales.}
  \label{fig:cell-phase-pointwise}
\end{figure}

For $n_{\ell,s}$ sampled elements $x_m$ of signal $s$ in layer $\ell$, compute
\begin{equation}
  \widehat c_{\ell,s,k}=\frac{1}{n_{\ell,s}}
  \sum_{m=1}^{n_{\ell,s}}
  \exp\!\left(2\pi i k\phi(x_m)\right),
  \qquad k=1,\ldots,16.
  \label{eq:cell-fourier}
\end{equation}
Summarize the first 16 frequencies by
\begin{equation}
  A_{\ell,s}=\left(\frac{1}{16}\sum_{k=1}^{16}
  |\widehat c_{\ell,s,k}|^2\right)^{1/2}.
  \label{eq:fourier-concentration}
\end{equation}
The statistic $A_{\ell,s}$ measures phase concentration over these frequencies.
We summarize layers by
\begin{equation}
  \widehat A_s
  =\operatorname{median}_{\ell} A_{\ell,s}.
  \label{eq:median-fourier-concentration}
\end{equation}
Figure~\ref{fig:cell-phase-overall} shows the distribution of $A_{\ell,s}$
across layers.  Figure~\ref{fig:cell-phase-pointwise} reports the layer medians
of $|\widehat c_{\ell,s,k}|$; for residual addition, the two residual sums'
magnitudes are averaged within each layer before taking the median.

Figure~\ref{fig:cell-phase-pointwise} shows consistent frequency patterns across
models.  RoPE products have peaks at $k=4,8,16$, and SiLU shows a similar
alternating pattern.  Residual sums have larger magnitudes at even frequencies,
with further increases at multiples of four and eight and the largest
magnitude at $k=16$.

\subsection{Rounding error and BF16 cell phase}
\label{app:rounding-error}

For the spacing $h(x)$ and phase $\phi(x)$ defined in
Section~\ref{app:cell-phase-fourier}, round-to-nearest-even conversion
$\Rbf$ gives, away from midpoint ties and overflow,
\begin{equation}
  \Rbf(x)-x=h(x)\rho(\phi(x)),
  \qquad
  \rho(u)=
  \begin{cases}
    -u, & 0\leq u<\tfrac12,\\
    1-u, & \tfrac12<u<1.
  \end{cases}
  \label{eq:bf16-rounding-error}
\end{equation}
At a midpoint, the parity of the adjacent BF16 significands determines the
rounding direction.  Away from ties, averaging $h(x)\rho(\phi(x))$ over the
joint distribution of cell widths and phases gives the mean rounding error.
Uniform phases independent of $h(x)$
give zero mean rounding error because $\int_0^1\rho(u)\,du=0$.
Exactly BF16-representable values also have zero rounding error, with
$\phi(x)=0$ and all Fourier coefficients equal to one.  The effect on subsequent
computations depends on their sensitivity to the rounded values.

\section{Additional Experiment on DAPO-Math-17k}
\label{app:dapo-math}

We repeat the four conditions in Section~\ref{sec:llm-rl-controls} with
DeepSeek-R1-Distill-Qwen-1.5B, replacing the training dataset with
DAPO-Math-17k~\citep{yu2025dapo}.  The PPO-clipped GRPO objective, sampling
configuration, optimization hyperparameters, and execution settings are the
same as for this model in Appendix~\ref{app:experimental-details} and
Table~\ref{tab:rl-training-details}; we also evaluate on AIME 2025.

\begin{figure}[htbp]
    \centering
    \includegraphics{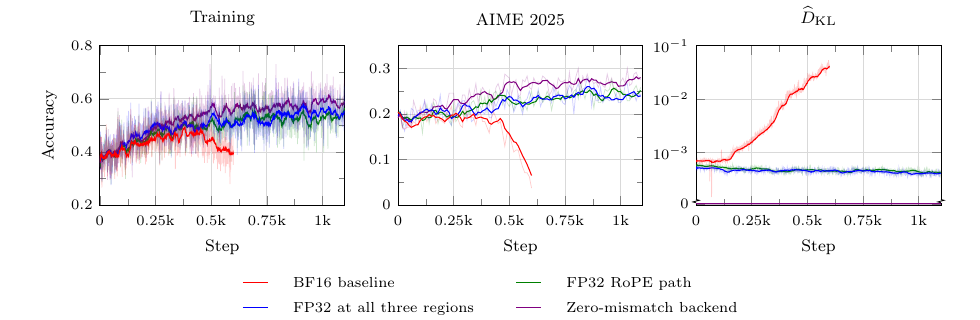}
    \caption{Training accuracy, AIME 2025 accuracy, and $\tokenkl$
    (Eq.~\ref{eq:per-token-kl}) for
    DeepSeek-R1-Distill-Qwen-1.5B trained on DAPO-Math-17k.
    Observations are smoothed using causal trailing means over
    20 observations for training accuracy and $\tokenkl$ and five observations
    for AIME 2025 accuracy.
    The BF16 baseline loses accuracy after initial improvement; both
    rounding-location alignments and the Zero-mismatch backend sustain
    training progress.}
    \label{fig:dapo-math-curves}
\end{figure}

Figure~\ref{fig:dapo-math-curves} shows training and AIME 2025 accuracy and
the inference--training discrepancy $\tokenkl$ (Eq.~\ref{eq:per-token-kl}).
Both targeted alignments and the Zero-mismatch backend sustain training
progress, whereas the BF16 baseline loses accuracy after initial
improvement.  The baseline's $\tokenkl$ rises sharply;
both targeted alignments retain small, nonzero $\tokenkl$, and the
Zero-mismatch backend remains at zero.  RoPE-path rounding-location alignment
thus also stabilizes training on this additional dataset.

\section{Training Overhead of Rounding-Location Alignment}
\label{app:training-efficiency}

We compare the BF16 baseline with FP32 intermediates along the RoPE path
and at all three regions (Section~\ref{sec:qwen-kernel-fusion}).  Experiments
use eight GPUs with fully sharded data parallelism (FSDP) in Hugging Face.
Each GPU processes one random 5,120-token sequence per step.  All conditions
use FP32 master weights and gradient reduction, BF16 compute parameters,
scaled dot-product attention, and non-reentrant gradient checkpointing;
compilation and TF32 are disabled.

Timing covers forward, full-vocabulary FP32 log-softmax and mean next-token
negative log likelihood over all prediction positions, and backward, including
communication.  Pretrained weights remain fixed; rollout, optimizer updates,
data loading, and gradient clearing are excluded.  Each run uses ten warmup
and twenty measured steps.
Step latency is the maximum wall-clock duration across ranks, measured after
a barrier with device synchronization before and after each timed step.
Table~\ref{tab:training-efficiency} reports latency changes averaged over
paired run ratios and peak allocated memory maximized over ranks and runs.

RoPE-path alignment adds 0.90--2.04\% latency with unchanged peak allocated
memory; three-region alignment adds 5.33--10.92\% latency and
0.41--0.89~GiB per GPU.

\begin{table}[!htbp]
  \centering
  \caption{Training-computation overhead.  Latency is the mean step
  latency; changes are
  relative to each model's BF16 baseline.}
  \label{tab:training-efficiency}
  \small
  \setlength{\tabcolsep}{4pt}
  \begin{tabular}{@{}llrrr@{}}
    \toprule
    Model & FP32 regions & \shortstack{Latency\\(ms/step)} &
    \shortstack{Latency\\change (\%)} &
    \shortstack{Memory\\(GiB/GPU)} \\
    \midrule
    \shortstack[l]{DeepSeek-R1-Distill-\\Qwen-1.5B}
      & None (baseline) & $702.61$ & --- & 12.62 \\
      & RoPE path & $713.76$ & $+1.59$ & 12.62 \\
      & All three & $779.31$ & $+10.92$ & 13.03 \\
    \midrule
    Qwen3-1.7B
      & None (baseline) & $878.04$ & --- & 12.36 \\
      & RoPE path & $895.96$ & $+2.04$ & 12.36 \\
      & All three & $946.70$ & $+7.82$ & 12.91 \\
    \midrule
    Qwen3-8B-Base
      & None (baseline) & $3025.63$ & --- & 19.00 \\
      & RoPE path & $3052.81$ & $+0.90$ & 19.00 \\
      & All three & $3186.77$ & $+5.33$ & 19.89 \\
    \bottomrule
  \end{tabular}
\end{table}

\section{Implementation of the Zero-Mismatch Backend}
\label{app:zero-mismatch-backend}

The Zero-mismatch backend in Section~\ref{sec:llm-rl-controls} uses the same
uncompiled Hugging Face forward implementation for inference and training,
preserving BF16 rounding locations.  We implement batch-invariant kernels
in Triton~\citep{tillet2019triton} for matrix multiplication, log-softmax,
mean reduction, and attention.  Our implementation is based on the code
released by \citet{he2025nondeterminism}.

The matrix and reduction kernels keep accumulation order independent of
batch size; log-softmax computes log probabilities in FP32, and mean
reduction covers normalization statistics.  Inference and training also
share the same attention kernel and key/value traversal order, with FP32
accumulation and BF16 outputs.  Cached decoding retains this reduction
scheme without separately reducing partitions of the key/value sequence,
aligning attention arithmetic with full-sequence recomputation.

\end{document}